\documentclass[runningheads]{llncs}

\usepackage[final,year=2026]{eccv}
\usepackage{eccvabbrv}
\usepackage{graphicx}
\usepackage{booktabs}
\usepackage{amsmath,amssymb}
\usepackage[accsupp]{axessibility}
\usepackage{hyperref}

\begin{document}

\title{Visual Autoregressive Priors for RAW-to-sRGB Image Signal Processing}
\titlerunning{VAR Priors for RAW-to-sRGB ISP}

\author{Tailai Chen\inst{1} \and
Xiaotong Luo\inst{2} \and
Yuan Gao\inst{1}\thanks{Corresponding author.} \and
Xin Jin\inst{1} \and
Wenjun Zeng\inst{1}}
\authorrunning{T. Chen et al.}
\institute{OmniVision-IDT Joint Laboratory for Intelligent Image Sensing,\\
Ningbo Key Laboratory of Spatial Intelligence and Digital Derivative,\\
Ningbo Institute of Digital Twin, Eastern Institute of Technology, Ningbo;\\
Zhejiang Key Laboratory of Industrial Intelligence and Digital Twin\\
\email{ygao@idt.eitech.edu.cn}
\and
The Hong Kong Polytechnic University}

\maketitle

\begin{abstract}
RAW-to-sRGB image signal processing (ISP) must recover perceptually faithful colors and fine details from sensor measurements, often under imperfect spatial alignment and missing camera metadata.
This paper presents, to the best of our knowledge, the first application of visual autoregressive (VAR) next-scale prediction over a discrete image codebook to the RAW-to-sRGB ISP task.
We adapt a frozen 1.10\,B-parameter VAR backbone for RAW-conditioned ISP with only 32.93\,M trainable parameters (2.99\%), and propose a frequency-decomposed color loss that separately supervises low-frequency tone via wavelet LL cosine similarity and chromatic edges via detail-band $\ell_1$.
On the Zurich RAW-to-sRGB benchmark, the method improves PSNR-Y from 21.31 to 21.89\,dB and reduces LPIPS from 0.276 to 0.218 on the full 1,204-image test set.
Diagnostic experiments show that the VAR prior preserves structure well, but continuous color transfer remains the dominant bottleneck: oracle affine correction recovers 3.8\,dB, while learned color heads yield marginal gains.
\end{abstract}

\section{Introduction}
\label{sec:intro}

Modern camera pipelines transform RAW sensor measurements into display-ready sRGB images through a cascade of operations: demosaicing, denoising, white balance, color correction, gamma mapping, and sharpening.
End-to-end neural ISP aims to replace this entire stack with a single learned mapping~\cite{ignatov2020replacing,ignatov2021learned}, but the task is considerably more challenging than standard image restoration.
In real paired datasets such as Zurich RAW-to-sRGB (ZRR), RAW inputs and sRGB targets are captured by different cameras with different optics, exposure behavior, and color rendering pipelines, and camera metadata may be absent entirely.
The model must therefore handle texture reconstruction, spatially misaligned supervision, and cross-camera color transfer simultaneously.

This challenge has shaped the neural ISP literature.
W-Net treats ZRR as a misaligned supervision problem with color-sensitive losses~\cite{kim2019wnet}.
LiteISPNet introduces a global color mapping module for alignment against a color-adjusted reference~\cite{zhang2021rawsrgb}.
AWNet and MW-ISPNet use wavelet and multi-scale components for detail preservation with enlarged receptive fields~\cite{dai2020awnet,ignatov2020aim}.
FourierISP separates phase (structure) and amplitude (style) in the frequency domain~\cite{he2024fourierisp}.
RMFA-Net argues that RAW-specific preprocessing directly affects color fidelity~\cite{li2024rmfa}.
These works establish a recurring principle: successful RAW-to-sRGB models need both spatial detail preservation and controlled color transfer.

Generative priors offer a complementary angle.
Diffusion models have been applied: DiffRAW conditions on LiteISPNet outputs~\cite{yi2024diffraw}, and ISPDiffuser separates grayscale reconstruction from histogram-guided color mapping~\cite{ren2025ispdiffuser}.
These report strong perceptual quality but require iterative sampling.
Visual autoregressive modeling (VAR)~\cite{tian2024var} generates images through next-scale prediction over vector-quantized latents, offering a coarse-to-fine prior naturally suited to ISP: the RAW image supplies scene content at all scales while the pretrained prior regularizes plausible sRGB structure.
However, a discrete codebook is not obviously suited to continuous camera operations---white balance, exposure compensation, and tone curves---that are fundamentally smooth, per-pixel transformations.

We therefore ask: \emph{can a frozen VAR prior be adapted for RAW-to-sRGB ISP with a small trainable conditioning branch, and where does this adaptation fail?}
Our approach freezes the entire VAR backbone and VQ-VAE, training only conditioning embeddings and cross-attention modules (2.99\% of total parameters).
To handle misaligned supervision, we propose a frequency-decomposed color loss: a Haar wavelet decomposes outputs and targets, and we apply cosine similarity on the LL subband for global tone while using $\ell_1$ on detail bands for chromatic edges.
This design draws on the structure--color separation from W-Net~\cite{kim2019wnet} and FourierISP~\cite{he2024fourierisp}, but operates in the wavelet domain for better spatial locality under misalignment.

On ZRR, the method improves PSNR-Y by 0.58\,dB and reduces LPIPS by 21\% over the baseline without color supervision.
However, diagnostic experiments reveal that remaining errors are dominated by brightness, contrast, and white-balance shifts rather than texture or structural artifacts.
Oracle per-image affine color correction recovers 3.8\,dB PSNR, demonstrating that the output already contains the necessary structural detail.
Yet learned color heads yield only marginal or negative gains, confirming that robust continuous color transfer under misaligned supervision remains the key open problem for discrete generative priors in ISP.

Our contributions are:
\begin{itemize}
    \item To the best of our knowledge, the first application of visual autoregressive modeling to RAW-to-sRGB ISP, with a parameter-efficient formulation adapting a frozen 1.10\,B-parameter VAR backbone using only 32.93\,M trainable parameters (2.99\%).
    \item A frequency-decomposed color loss using wavelet-domain cosine similarity for tone and $\ell_1$ for chromatic edges, robust to spatial misalignment in cross-camera datasets.
    \item Diagnostic experiments demonstrating that the VAR prior captures structure well, while continuous color transfer remains the main limitation, with oracle affine correction recovering 3.8\,dB PSNR.
\end{itemize}

\section{Related Work}
\label{sec:related}

\paragraph{Learned RAW-to-sRGB ISP.}
PyNET and the ZRR benchmark introduced a practical paired RAW--DSLR evaluation setting~\cite{ignatov2020replacing}, while the Mobile AI 2021 challenge emphasized mobile ISP under resource constraints~\cite{ignatov2021learned}.
W-Net~\cite{kim2019wnet} addresses misalignment with a two-stage U-Net and color-sensitive loss.
AWNet~\cite{dai2020awnet} uses wavelet-domain processing and global context, while MW-ISPNet~\cite{ignatov2020aim} employs multi-level wavelet components.
LiteISPNet~\cite{zhang2021rawsrgb} models geometric correspondence explicitly for learning with inaccurately aligned supervision.
FourierISP~\cite{he2024fourierisp} decouples structure and style via phase and amplitude separation.
RMFA-Net~\cite{li2024rmfa} designs RAW-specific preprocessing to preserve color fidelity.
Our work takes a different approach: probing how far a large pretrained discrete generative prior can be pushed for ISP via parameter-efficient adaptation.

\paragraph{Generative priors for ISP.}
DiffRAW~\cite{yi2024diffraw} conditions a diffusion model on LiteISPNet outputs to enhance perceptual quality.
ISPDiffuser~\cite{ren2025ispdiffuser} separates grayscale detail from histogram-guided color consistency, achieving strong ZRR results at higher iterative inference cost.
Our work explores a different class of generative prior: a frozen next-scale autoregressive transformer with deterministic single-pass inference.

\paragraph{Discrete visual priors and VAR for low-level vision.}
Vector-quantized autoencoders~\cite{oord2017vqvae} represent images through a discrete codebook and serve as a common substrate for generative modeling.
VAR~\cite{tian2024var} redefines autoregression as next-scale prediction, enabling coarse-to-fine generation over VQ latents that naturally captures the hierarchical structure of natural images.
Very recently, this paradigm has been extended to low-level vision: VARSR~\cite{qu2025varsr} adapts VAR for image super-resolution with prefix tokens and a diffusion refiner, and RestoreVAR~\cite{rajagopalan2025restorevar} extends VAR to all-in-one image restoration via cross-attention conditioning on degraded image latents.
To the best of our knowledge, our work is the \emph{first} to apply the VAR framework to the RAW-to-sRGB ISP task, which presents unique challenges of cross-camera color transfer and spatially misaligned supervision not encountered in standard super-resolution or restoration settings.

\paragraph{Color correction under misalignment.}
Color is central to ISP but entangled with exposure, white balance, tone curves, and the target camera's rendering style.
In misaligned datasets like ZRR, direct pixel-level supervision can degrade high-frequency details or encourage desaturated colors.
Prior work addresses this through color-aware objectives~\cite{kim2019wnet}, global context modules~\cite{dai2020awnet,zhang2021rawsrgb}, frequency-domain style separation~\cite{he2024fourierisp}, histogram-guided colorization~\cite{ren2025ispdiffuser}, and sensor-aware tone modeling~\cite{li2024rmfa}.
Our experiments separate the VAR branch's discrete detail generation from frequency-domain color objectives and post-hoc color heads to isolate the chromatic mismatch contribution.

\section{Method}
\label{sec:method}

\begin{figure}[!htb]
    \centering
    \includegraphics[width=0.9\linewidth]{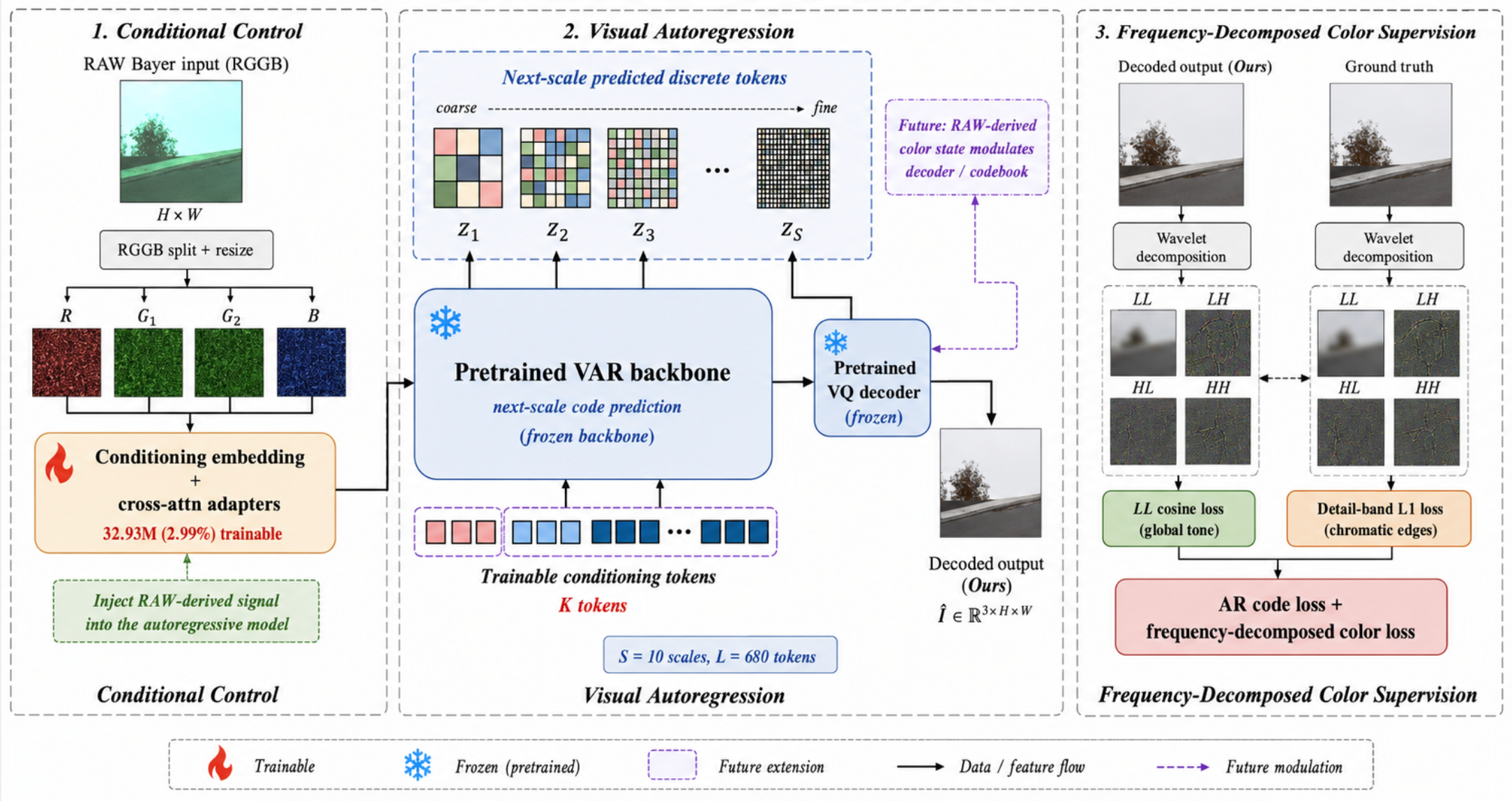}
    \caption{Overview of the proposed RAW-to-sRGB formulation. The frozen VQ-VAE tokenizes the target into multi-scale codes for training supervision. The frozen VAR transformer predicts codes conditioned on a RAW-derived signal through trainable adapters. A frequency-decomposed color loss supervises tone and chromatic edges in the wavelet domain.}
    \label{fig:framework}
\end{figure}

Figure~\ref{fig:framework} illustrates the proposed framework.
The pipeline consists of four stages: VQ-VAE tokenization, RAW conditioning, autoregressive code prediction with a frozen VAR backbone, and frequency-decomposed color supervision.
During inference, the RAW condition guides the frozen model to predict multi-scale VQ codes, decoded into the final sRGB output in a single deterministic forward pass.

\subsection{VQ-VAE Tokenization}
\label{sec:vqvae}

The foundation of our approach is a pretrained VQ-VAE~\cite{oord2017vqvae} that maps continuous images into discrete code sequences.
The encoder maps an input $x \in \mathbb{R}^{H \times W \times 3}$ to a spatial feature map, quantized element-wise against a learned codebook $\mathcal{C} = \{e_k\}_{k=1}^{K}$ with $K = 4096$ entries and latent dimension $C_{\mathrm{vae}} = 32$.
Following VAR~\cite{tian2024var}, the VQ-VAE produces $S = 10$ scale levels with resolutions $\{1{\times}1, 2{\times}2, \ldots, 10{\times}10\}$, yielding $L = 680$ total tokens per image.
The coarsest scale captures global color and layout, while finer scales add local detail and texture.

This multi-scale representation implies a fidelity \emph{ceiling}: the VQ-VAE reconstruction defines the best any downstream autoregressive prediction can achieve.
If the codebook cannot faithfully represent the target camera's color gamut, the system inherits this limitation regardless of conditioning quality.
We analyze this ceiling in Sec.~\ref{sec:vae-ceiling}.

\subsection{RAW Conditioning}
\label{sec:raw-cond}

Given a RAW Bayer image $r \in \mathbb{R}^{H_r \times W_r}$ with an RGGB color filter array, we construct a three-channel conditioning image $c \in \mathbb{R}^{256 \times 256 \times 3}$ by splitting the mosaic into R, G, B channels (averaging the two green pixels), then bilinearly resizing to $256 \times 256$ and normalizing to $[-1, 1]$.
This packing prioritizes compatibility with the pretrained three-channel conditioning embedding.
The two green pixels carry slightly different spatial information due to their Bayer grid offset; a more sophisticated CFA-aware packing could preserve this, but we treat it as secondary relative to the color transfer problem.

\subsection{Parameter-Efficient Adaptation}
\label{sec:peft}

The target sRGB image $x$ is tokenized by the frozen VQ-VAE encoder into multi-scale discrete code maps $\{z_s\}_{s=1}^{S}$.
The VAR transformer predicts these codes autoregressively from coarse to fine, conditioned on $c$ via cross-attention.
At each scale $s$, the model receives all coarser-scale codes $z_{<s}$ and predicts $z_s$ in parallel across spatial positions.

We freeze all pretrained VAR parameters $\theta$ and train only the conditioning parameters $\phi$ (conditioning embedding and cross-attention projections):
\begin{equation}
    \mathcal{L}_{\mathrm{AR}}
    = - \sum_{s=1}^{S} \sum_{i} \log p_{\theta,\phi}(z_{s,i} \mid z_{<s}, z_{s,<i}, c),
\end{equation}
where $z_{s,i}$ is the target code at scale $s$ and position $i$.
This yields 32.93\,M trainable parameters out of 1.10\,B total (2.99\%), preserving the pretrained visual prior while adapting to RAW-conditioned generation.
The frozen VQ-VAE contributes an additional 108.95\,M non-updated parameters.

\subsection{Frequency-Decomposed Color Supervision}
\label{sec:color-loss}

Direct pixel-level losses between decoded output and DSLR target are problematic in ZRR due to residual spatial misalignment.
Small spatial offsets create large pixel gradients that pull the model toward blurry, spatially averaged outputs.
We therefore decompose supervision in the wavelet domain, which provides spatial locality and a natural separation between low-frequency color and mid-frequency edges.

Given prediction $\hat{x}$ and target $x$, a single-level Haar wavelet yields four subbands: approximation (LL) and detail (LH, HL, HH).
We apply cosine similarity on LL for global tone and white balance:
\begin{equation}
    \mathcal{L}_{\mathrm{LL}}
    = 1 - \frac{\langle \mathrm{LL}(\hat{x}),\; \mathrm{LL}(x) \rangle}
    {\|\mathrm{LL}(\hat{x})\|_2\,\|\mathrm{LL}(x)\|_2 + \epsilon},
    \label{eq:color-loss}
\end{equation}
and $\ell_1$ on detail subbands for chromatic edges:
\begin{equation}
    \mathcal{L}_{\mathrm{detail}}
    = \frac{\mu}{3}\sum_{d \in \{\text{LH, HL, HH}\}}
    \|d(\hat{x}) - d(x)\|_1,
    \label{eq:freq-loss}
\end{equation}
with $\mu = 0.1$ to down-weight details relative to the LL tone loss.

The cosine similarity in $\mathcal{L}_{\mathrm{LL}}$ is invariant to global brightness scaling, important because the smartphone and DSLR may have substantially different exposure levels.
The LL downsampling absorbs small spatial offsets that would cause large errors in a full-resolution pixel loss, while the detail-band $\ell_1$ encourages preservation of chromatic edge structure without requiring exact alignment.

The total training objective is:
\begin{equation}
    \mathcal{L} = \mathcal{L}_{\mathrm{AR}}
    + \lambda_{\mathrm{freq}}(\mathcal{L}_{\mathrm{LL}} + \mathcal{L}_{\mathrm{detail}}),
    \label{eq:total-loss}
\end{equation}
with $\lambda_{\mathrm{freq}} = 0.005$, balancing discrete code prediction with continuous color supervision.

\paragraph{Differentiable decoding during training.}
The image-domain loss does not backpropagate through an argmax operation.
For each token logit vector $\ell_{s,i}$, we compute a soft assignment
$p_{s,i}=\operatorname{softmax}(\ell_{s,i})$ and its expected codebook embedding
$\bar{e}_{s,i}=p_{s,i}\mathcal{C}$.
The expected embeddings from all scales are accumulated by the frozen residual-quantization path and decoded by the frozen VQ-VAE to obtain $\hat{x}$.
Consequently, gradients from the wavelet losses pass through the decoder, the fixed codebook multiplication, and the softmax to the predicted logits and trainable conditioning parameters $\phi$; the VAR backbone, codebook, and VQ-VAE remain frozen.
At inference, we replace the soft assignments with deterministic top-1 code selection.

\paragraph{Comparison with Fourier-domain approaches.}
Our wavelet decomposition shares conceptual ground with FourierISP~\cite{he2024fourierisp}, which separates phase (structure) and amplitude (style) globally.
The key advantage of the wavelet approach is spatial locality: each coefficient reflects a local neighborhood rather than a global frequency bin.
Under misalignment, a small shift causes bounded perturbations in local wavelet coefficients but large phase errors across the Fourier spectrum, making wavelet supervision more robust for cross-camera datasets.

\subsection{Post-Hoc Color Diagnostics}
\label{sec:posthoc}

To quantify how much of the remaining gap is attributable to color rather than structural errors, we evaluate three lightweight post-hoc color modules on the frozen model output $y$, intentionally constrained to prevent texture hallucination:

\begin{itemize}
    \item \textbf{Oracle affine}: A per-image $3 \times 4$ affine matrix fit by least-squares from output to ground truth. Since it uses test-time ground truth, it is not deployable---it provides an \emph{upper bound} on linear color-only correction.

    \item \textbf{Low-frequency residual head}: A small CNN with average-pooling bottleneck predicts a spatially smooth residual $\Delta_{\mathrm{low}}$ from the concatenation of $y$ and $c$, producing $y' = y + \Delta_{\mathrm{low}}(y, c)$.

    \item \textbf{Adaptive CCM}: A lightweight network predicts a per-image $3 \times 3$ matrix $A$ and bias $b$ from global average-pooled features, producing $y' = Ay + b$, trained with oracle-distilled supervision.
\end{itemize}

The contrast between oracle upper bound and learned modules reveals whether the bottleneck is structural or chromatic.

\section{Experiments}
\label{sec:experiments}

We first describe the experimental setup, then present quantitative and qualitative results, analyze the VQ-VAE reconstruction ceiling, and finally conduct post-hoc color correction diagnostics to isolate the structure--color bottleneck.

\subsection{Setup}

\paragraph{Dataset.}
We evaluate on the Zurich RAW-to-sRGB (ZRR) dataset~\cite{ignatov2020replacing}, pairing Huawei P20 Pro smartphone RAW captures with Canon 5D Mark IV DSLR images.
The training set contains 46,839 paired patches ($448 \times 448$), and the test set contains 1,204 patches.
Due to different lens assemblies, sensor sizes, and color pipelines, there is inherent spatial misalignment and substantial color mismatch between pairs, making ZRR particularly challenging.

\paragraph{Metrics.}
For all of our ablations, we report PSNR and SSIM on the luminance (Y) channel using one fixed evaluation script; this reduces the impact of cross-camera color shifts and enables controlled within-method comparisons.
We also report LPIPS~\cite{zhang2018perceptual} for perceptual similarity and secondary no-reference scores CLIPIQA and MUSIQ, though we emphasize full-reference fidelity because no-reference metrics can favor aesthetically pleasing but unfaithful reconstructions.

\paragraph{Architecture and training.}
The VAR backbone has depth 24, embedding dimension 1536, and 24 attention heads (1.10\,B parameters).
The VQ-VAE uses $K = 4096$ codes with $C_{\mathrm{vae}} = 32$, producing $S = 10$ scales with $L = 680$ tokens (108.95\,M parameters).
Both are entirely frozen; only conditioning embedding and cross-attention projections are trained (32.93\,M, 2.99\%).
We train with AdamW~\cite{loshchilov2019adamw} ($\mathrm{lr} = 10^{-4}$, cosine annealing).
At inference, deterministic top-$k{=}1$ decoding produces a single output in one forward pass.

\subsection{Main Results}

\begin{table}[!htb]
\centering
\small
\setlength{\tabcolsep}{4pt}
\caption{Ablation on the full ZRR test set (1,204 images). Each row adds one component to the baseline.}
\label{tab:zrr-main}
\begin{tabular}{lccccc}
\toprule
Configuration & PSNR-Y\,$\uparrow$ & SSIM-Y\,$\uparrow$ & LPIPS\,$\downarrow$ & CLIPIQA\,$\uparrow$ & MUSIQ\,$\uparrow$ \\
\midrule
Baseline & 21.312 & 0.7272 & 0.2764 & 0.4538 & 47.80 \\
+ Spatial color loss & 21.681 & 0.7356 & 0.2543 & 0.4229 & 44.97 \\
+ Freq-decomposed loss & \textbf{21.891} & \textbf{0.7562} & \textbf{0.2176} & 0.4107 & 43.52 \\
\bottomrule
\end{tabular}
\end{table}

Table~\ref{tab:zrr-main} presents the ablation.
The baseline with only $\mathcal{L}_{\mathrm{AR}}$ achieves 21.312\,dB PSNR-Y.
Adding spatial color loss improves PSNR-Y by 0.37\,dB; replacing it with the frequency-decomposed loss yields 21.891\,dB (+0.58\,dB total), SSIM-Y 0.7562, and LPIPS 0.2176 (21.3\% reduction).
The monotonic decrease in no-reference scores is expected: stronger DSLR-target supervision pulls outputs from generic aesthetic preferences---an acceptable trade-off for faithful reconstruction.

\begin{table}[!htb]
\centering
\caption{Published ZRR landscape under nonidentical evaluation protocols. Prior-work values are RGB-channel results quoted from the corresponding papers, whereas ours is evaluated on the Y channel over the full test set. The two blocks provide context and must not be numerically ranked across protocols.}
\label{tab:zrr-landscape}
\begin{tabular}{lcccc}
\toprule
Method & Evaluation & PSNR\,$\uparrow$ & SSIM\,$\uparrow$ & LPIPS\,$\downarrow$ \\
\midrule
PyNet~\cite{ignatov2020replacing} & RGB, published & 21.19 & 0.747 & 0.193 \\
AWNet-R~\cite{dai2020awnet} & RGB, published & 21.42 & 0.748 & 0.198 \\
AWNet-D~\cite{dai2020awnet} & RGB, published & 21.53 & 0.749 & 0.212 \\
MW-ISPNet~\cite{ignatov2020aim} & RGB, published & 21.42 & 0.754 & 0.213 \\
LiteISPNet~\cite{zhang2021rawsrgb} & RGB, published & 21.55 & 0.749 & 0.187 \\
FourierISP~\cite{he2024fourierisp} & RGB, published & 21.65 & 0.755 & 0.182 \\
DiffRAW~\cite{yi2024diffraw} & RGB, published & 21.31 & 0.743 & 0.145 \\
ISPDiffuser~\cite{ren2025ispdiffuser} & RGB, published & 21.77 & 0.754 & 0.157 \\
\midrule
Ours & Y, ours & 21.89 & 0.756 & 0.218 \\
\bottomrule
\end{tabular}
\end{table}

Table~\ref{tab:zrr-landscape} provides landscape context rather than a head-to-head ranking.
Because the published baselines use RGB-channel PSNR/SSIM while our retained evaluation uses the Y channel, differences between the two blocks are not directly attributable to model quality.
Our quantitative claims therefore rest on the matched-protocol ablation in Table~\ref{tab:zrr-main}; a unified RGB re-evaluation of all methods remains necessary for strict comparison.
The published LPIPS values nevertheless expose the main practical gap: our LPIPS (0.218) trails FourierISP (0.182) and diffusion methods (DiffRAW: 0.145; ISPDiffuser: 0.157), consistent with the residual color and tone errors observed qualitatively.

\subsection{Qualitative Analysis}

\begin{figure}[!htb]
    \centering
    \includegraphics[width=0.75\linewidth]{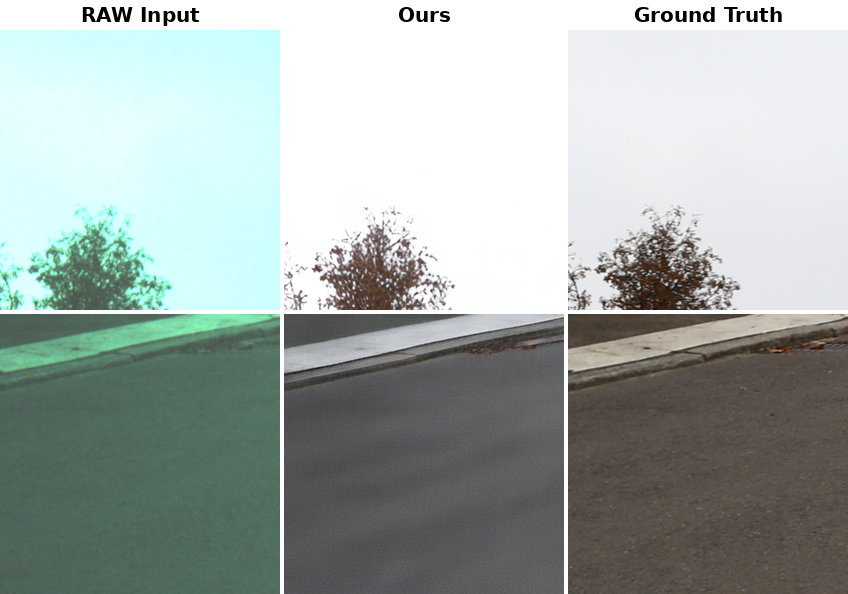}
    \caption{Qualitative results on ZRR. Each row: RAW input (left), our output (middle), ground truth (right). The model preserves structure and edges; residual errors are global color, brightness, and contrast shifts.}
    \label{fig:zrr-teaser}
\end{figure}

Figure~\ref{fig:zrr-teaser} shows representative outputs.
The model recovers plausible scene structure and edge detail from heavily color-shifted RAW inputs.
Object boundaries, texture patterns, and spatial layout are faithfully preserved.
However, remaining errors are predominantly \emph{global}: outputs often appear too gray, too bright, or with lower contrast than the DSLR target---characteristic of white-balance and tone-curve mismatches rather than texture or structural failures.
This visual pattern motivates the color correction diagnostics below.

\subsection{VQ-VAE Reconstruction Ceiling}
\label{sec:vae-ceiling}

\begin{figure}[!htb]
    \centering
    \includegraphics[width=0.65\linewidth]{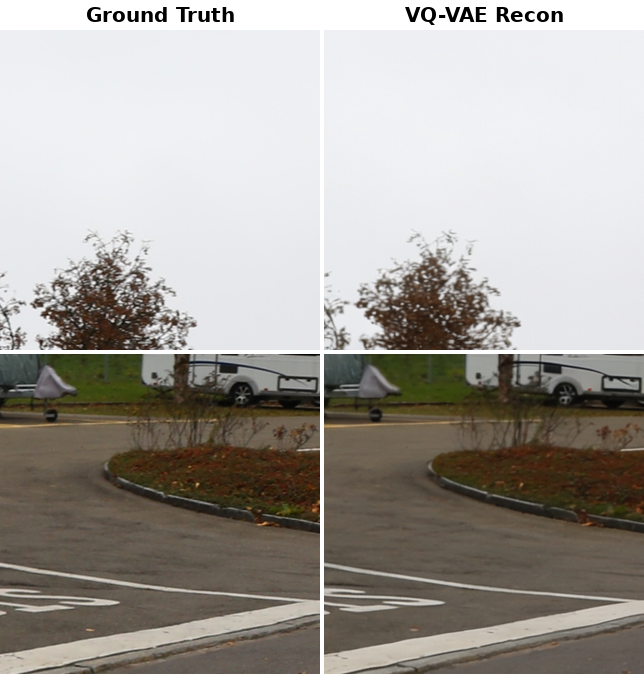}
    \caption{VQ-VAE reconstruction ceiling. Each row: ground truth (left), VQ-VAE encode-decode reconstruction (right). The codebook preserves structure, color, and most detail; artifacts appear only in smooth gradient regions.}
    \label{fig:vae-ceiling}
\end{figure}

Before analyzing the autoregressive model's errors, we characterize the VQ-VAE fidelity ceiling.
Figure~\ref{fig:vae-ceiling} compares ground-truth ZRR targets with their VQ-VAE encode-decode reconstructions (bypassing the VAR transformer).
The frozen VQ-VAE preserves scene layout, boundaries, and color with high fidelity.
Quantization artifacts appear primarily in smooth gradients (sky, walls, defocused backgrounds) where the discrete codebook cannot represent every tonal variation.
These artifacts are substantially smaller than errors from autoregressive prediction, confirming that the $K = 4096$ codebook is not the bottleneck---the dominant error source is RAW-conditioned code selection.

\subsection{Color Correction Diagnostics}

\begin{table}[!htb]
\centering
\caption{Post-hoc color diagnostics on a 100-image ZRR subset. The oracle affine uses each test target to fit a per-image affine transform and is not deployable; it estimates the ceiling of color-only correction.}
\label{tab:color-heads}
\begin{tabular}{lccc}
\toprule
Setting & PSNR-Y\,$\uparrow$ & SSIM-Y\,$\uparrow$ & LPIPS\,$\downarrow$ \\
\midrule
Oracle affine, before & 21.129 & 0.7123 & 0.2892 \\
Oracle affine, after & \textbf{24.950} & \textbf{0.7269} & \textbf{0.2815} \\
\midrule
Low-freq residual, before & 21.681 & 0.7357 & 0.2543 \\
Low-freq residual, after & 21.694 & 0.7357 & 0.2543 \\
\midrule
Adaptive CCM, before & 21.678 & 0.7356 & 0.2543 \\
Adaptive CCM, after & 21.391 & 0.7359 & 0.2648 \\
\bottomrule
\end{tabular}
\end{table}

Table~\ref{tab:color-heads} presents the central diagnostic result.
The oracle affine raises PSNR-Y by 3.8\,dB (21.13 to 24.95), confirming that the output already contains the structural detail for high-fidelity reconstruction---the missing ingredient is correct color and tone mapping.

Neither learned head realizes this potential.
The low-frequency residual head provides only 0.013\,dB, suggesting spatially smooth additive correction is insufficient.
The adaptive CCM actually \emph{worsens} LPIPS from 0.254 to 0.265, indicating overcorrection.
We attribute this to: (1) substantial per-image variation in exposure, white balance, and illumination in ZRR; (2) misalignment limiting supervisory signal for precise color correspondences; and (3) teacher matrices computed against misaligned targets introducing systematic noise.

\begin{figure}[!htb]
    \centering
    \includegraphics[width=0.85\linewidth]{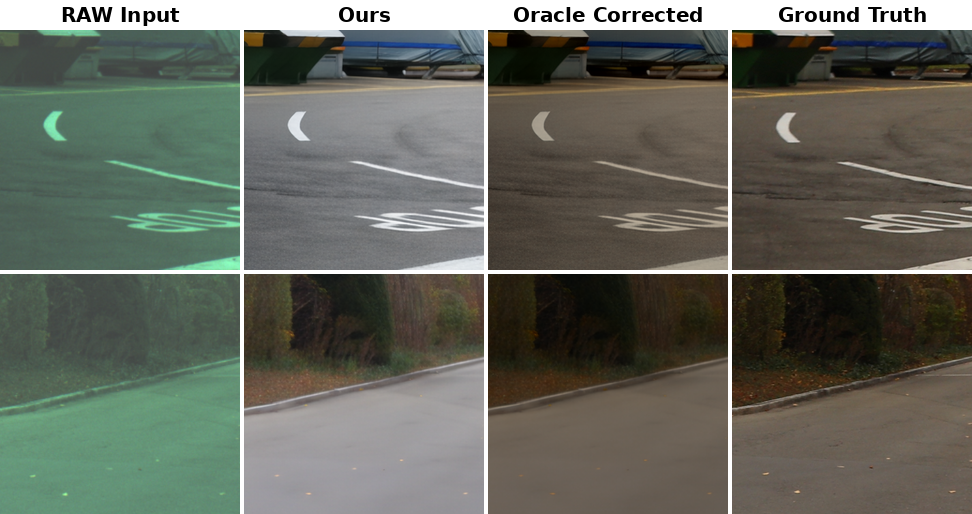}
    \caption{Oracle affine correction. Each row, left to right: RAW input, model output, oracle-corrected output, ground truth. The oracle recovers target color and contrast, confirming the output contains necessary structural detail.}
    \label{fig:color-comparison}
\end{figure}

Figure~\ref{fig:color-comparison} provides visual evidence.
The model output captures scene content faithfully but exhibits color and brightness shift.
After oracle correction, the output closely matches the ground truth, confirming that the VAR prior has solved structure reconstruction and the remaining gap is a learnable but currently unsolved color transform.

\section{Discussion}
\label{sec:discussion}

Our experiments demonstrate that VAR priors effectively capture image structure for ISP, producing outputs with coherent edges, textures, and layout.
However, the discrete codebook should not be expected to simultaneously solve all continuous camera operations through code selection alone.
This echoes the design principle shared by recent high-performing methods: FourierISP~\cite{he2024fourierisp} separates structure-related phase from color-related amplitude, and ISPDiffuser~\cite{ren2025ispdiffuser} assigns diffusion to grayscale texture with a separate module for color consistency.

The 3.8\,dB oracle gap (Table~\ref{tab:color-heads}) represents a substantial opportunity.
A natural next step is a \emph{color-modulated} architecture in which a compact color state is predicted from RAW statistics and injected into the VQ decoder or VAR attention blocks through adaptive normalization, cross-attention bias, or low-rank codebook offsets.
This keeps the discrete prior responsible for geometry and texture while providing a continuous path for exposure, white balance, and target-camera style.
The failure of post-hoc correction (Table~\ref{tab:color-heads}) suggests this modulation should be trained jointly with the autoregressive objective, rather than applied after quantization and decoding.

\paragraph{Limitations.}
The model operates at $256{\times}256$, below the native resolution of many practical ISP applications; scaling would require tiling with overlap blending or a resolution-adapted VAR architecture.
The frozen codebook was trained on ImageNet and may not optimally represent specific camera color gamuts; fine-tuning or expanding the codebook for ISP is a promising direction.
We evaluate on a single benchmark (ZRR); generalization to other sensor--target pairs remains to be verified.
Moreover, Table~\ref{tab:zrr-landscape} combines results reported under different channel protocols and is contextual rather than a strict ranking; a unified evaluation is required for direct comparison.
The 1.10\,B-parameter backbone is impractical for mobile deployment, though it serves our diagnostic purpose.

\section{Conclusion}
\label{sec:conclusion}

We presented the first study of visual autoregressive priors for RAW-to-sRGB ISP.
A frozen 1.10\,B-parameter VAR backbone, adapted with only 32.93\,M trainable parameters via conditioning embeddings and cross-attention, produces detailed sRGB reconstructions on ZRR.
The frequency-decomposed color loss improves full-test fidelity to 21.89\,dB PSNR-Y and 0.218 LPIPS.
Oracle and learned color-correction experiments confirm that the remaining bottleneck is continuous color transfer rather than detail synthesis, with oracle affine correction recovering 3.8\,dB.
These findings motivate future VAR-based ISP architectures combining discrete structure generation with jointly trained continuous color modulation.

\section*{Acknowledgments}
This work was supported by the OmniVision-IDT Joint Laboratory for Intelligent Image Sensing under the project ``Research on Frontier Technologies of Intelligent Image Sensing''. The computing for this research was supported by High Performance Computing Platform at Eastern Institute of Technology, Ningbo.

\clearpage
\bibliographystyle{splncs04}
\bibliography{main}

\end{document}